\documentclass{ifacconf}

\usepackage{graphicx}      
\usepackage{natbib}        
\usepackage{amsfonts}
\usepackage{amsmath}
\usepackage{xcolor,soul}
\usepackage{todonotes}
\usepackage{multirow}
\usepackage{url}

\newtheorem{theorem}{Theorem}
\newtheorem{proposition}{Proposition}
\newif\ifnotes\notestrue

\def\hsteph#1{} % switch to h to hide the direct comments
\newcommand{\R}{\ensuremath{\mathbb{R}}}

\begin{document}
\begin{frontmatter}

\title{Classifying histograms of medical data using information geometry of beta distributions} 

\thanks[footnoteinfo]{Sponsor and financial support acknowledgment
goes here. Paper titles should be written in uppercase and lowercase
letters, not all uppercase.}

\author[First]{Alice Le Brigant} 
\author[Second]{Nicolas Guigui} 
\author[Third,Fourth]{Sana Rebbah}
\author[Third]{Stéphane Puechmorel}

\address[First]{Universit\'{e} Paris 1 Panth\'{e}on Sorbonne, 
   Paris, France} %(e-mail: alice.le-brigant@univ-paris1.fr)}
\address[Second]{Universit\'{e} C\^{o}te d'Azur, Inria Epione, 
   Sophia Antipolis, France} %(e-mail: nicolas.guigui@inria.fr).}
\address[Third]{Ecole Nationale de l'Aviation Civile, 
   Toulouse, France} %(e-mail: stephane.puechmorel@enac.fr)}
\address[Fourth]{Toulouse NeuroImaging Center, Université de Toulouse, Inserm, UPS, France}

\begin{abstract}                % Abstract of not more than 250 words.
In this paper, we use tools of information geometry to compare, 
average and classify histograms. Beta distributions are fitted to the histograms and the corresponding Fisher information geometry is used for comparison.
We show that this geometry is negatively curved, which guarantees
uniqueness of the notion of mean, and makes it suitable to
%constitutes an argument in favor of its use to 
classify histograms through the popular K-means algorithm. We illustrate the use of these geometric tools in supervised and unsupervised classification procedures of two medical data-sets,
%in medical applications: the study of 
cardiac shape deformations for the detection of pulmonary hypertension and brain cortical thickness for the diagnosis of Alzheimer's disease.
\end{abstract}


\end{frontmatter}
%===============================================================================

\section{Introduction}

The differential geometric approach to probability theory and statistics has met increasing interest in the past years, from the theoretical point of view as well as in applications. In this approach, probability distributions are seen as elements of a differentiable manifold, on which a metric structure is defined through the choice of a Riemannian metric. Two very important ones are the Wasserstein metric, central in optimal transport, and the Fisher information metric (also called Fisher-Rao metric), essential in information geometry. Unlike optimal transport, information geometry is foremost concerned with parametric families of probability distributions, and defines a Riemannian structure on the parameter space using the Fisher information matrix, see \cite{fisher1922}. The Fisher information is the Hessian of the Kullback-Leibler divergence, a popular measure of dissimilarity between probability distributions that does not verify the properties of a distance. In parameter estimation, it can be interpreted as the quantity of information contained in the model about unknown parameters. %As the Hessian of the well-known Kullback-Leibler divergence, it measures the capacity to distinguish between two different values of the parameter through the notion of curvature. 
\cite{rao1992} showed that it could be used to locally define a scalar product on the space of parameters, interpreted as a Riemannian metric.

An important feature of this metric is that it is invariant under any diffeomorphic change of parameterization. Indeed, it is induced by a more general Riemannian metric on the infinite-dimensional space of smooth, non parametric probability densities, and therefore it is invariant to coordinate change; see e.g. \cite{friedrich1991}. It can be seen as a natural choice of metric as it is the only Riemannian metric that is invariant with respect to transformation by a sufficient statistic, or a diffeomorphic transformation of the support in the non-parametric case \citep{cencov2000, bauer2016}.
%In fact, considering the infinite-dimensional space of probability densities on a given manifold $M$, there is a unique metric, which also goes by the name Fisher-Rao, that is invariant with respect to the action of the diffeomorphism group of $M$; see \cite{cencov2000}, \cite{bauer2016}. 
%This metric induces the regular Fisher information metric on the finite dimensional submanifolds corresponding to the parameterized statistical models of interest in information geometry. 
Arguably the most famous example of Fisher information geometry of a statistical model is that of the univariate Gaussian model, which is hyperbolic. The geometries of other parametric families such as the multivariate Gaussian model \citep{atkinson1981,skovgaard1984}, the family of gamma distributions \citep{arwini2008, rebbah2019geometry}, or more generally location-scale models \citep{said2019}, among others, have also received a lot of attention. %The case of the Gaussian model was extensively studied in \cite{} along with other parameterized families, and

In this work, we focus on beta distributions, a family of probability measures on $[0,1]$ used to model random variables defined on a compact interval in a wide variety of applications, e.g. in Bayesian inference as conjugate prior for several discrete probability distributions \citep{o1999bayesian}, or to model percentages and proportions in genomic studies \citep{yang2017beta}. Up to our knowledge, the information geometry of beta distributions has not yet received much attention. In Section \ref{sec:theory}, we give new results and properties for this geometry by deriving the metric matrix, and the geodesic equations. We compute the sectional curvature and show it has negative sign everywhere. This result is of particular interest to ensure the existence of Fréchet means. By numerically solving the geodesic equation, we may compute geodesic distances and means between probability distributions.

In Section \ref{sec:applications}, we exemplify the use of this geometry in the analysis of histograms extracted from segmented medical images. Traditional statistical methods are challenged in medical data analysis as data-sets often have very few samples ($\sim 10 -  100$) whereas observations are very high dimensional ($\sim 10^4 - 10^6$) and missing data are ubiquitous. Moreover anatomical measurements are often obtained through complex processing pipelines that result in various numbers of features (e.g. mesh representations of organs' surface may have different numbers of vertices), resulting in the lack of a common representation space for the data. 

To bypass these issues, it is thus helpful to consider histograms across the whole region of interest and to represent them in the space of beta distributions. The obtained representation has only two dimensions which is useful for visualization and interpretation of the results. We demonstrates that classification in this space achieves similar accuracy in discriminating patients from healthy controls as in the high-dimensional data space.

%The paper is organized in two parts. Section~\ref{sec:theory} gives theoretical results on the Fisher information geometry of beta distributions, while Section~\ref{sec:applications} is devoted to the medical applications.

\section{Fisher geometry of beta distributions}\label{sec:theory}

\subsection{The Fisher information metric}

Let $\Theta$ be an open set of $\R^d$ and let $\mu$ be a measure defined on a measurable space $(E, \mathcal{T})$. A parameterized family of distributions on the parameter space $\Theta$ is a set of probability measures absolutely continuous with respect to $\mu$, that is:
\begin{equation*}
\mathcal{P}_\Theta=\{p(\cdot;\theta) \mu, \theta\in \Theta\}.
\end{equation*}
When the mapping $\theta\in\Theta \mapsto p(x, \theta)$ is differentiable for $\mu$-almost all $x \in E$ and the model satisfies certain regularity conditions, the Fisher information matrix at $\theta$ is defined to be:
\begin{equation*}
% attention au signe dans l'expression ci-desssous (j'ai rajouté le -)
%I(\theta) = - \left[E\left( \frac{\partial^2}{\partial \theta_i\partial\theta_j} \ln p(X;\theta)\right)\right]_{1\leq i,j\leq d}.
I(\theta) = \left[E \partial_i l(x,\theta)\partial_j l(x, \theta) \right]_{1\leq i,j\leq d},
\end{equation*}
where $l(x,\theta)=\log p(x,\theta)$.
As an open subset of $\mathbb{R}^d$, $\Theta$ is a differentiable manifold with $T_\theta\Theta \simeq \mathbb{R}^d$. The Fisher information defines a Riemannian metric on $\Theta$ called the Fisher information metric%endows it with a Riemannian structure:
%can be equipped with a Riemannian metric using this quantity. This gives the Fisher information metric on the parameter space %$\Theta$, which defines the scalar product of two tangent vectors $u,v\in T_\theta\Theta \simeq \mathbb{R}^d$ at point $\theta\in\Theta$ as
\begin{equation*}
%ds^2 = I(\theta)_{ij}d\theta^id\theta^j,
(u,v) \in T_\theta \Theta \mapsto g_\theta(u,v) = u^t I(\theta)v,
\end{equation*}
where $u^t$ denotes the transpose of the vector $u$. 
%By extension, we talk of the Fisher geometry of the parameterized family $\mathcal P_\Theta$, and of the Riemannian manifold %$(\mathcal P_\Theta, g)$.
By a slight abuse of language, we will talk of the geometry of $\mathcal{P}_\Theta$, implicitly referring to the metric structure induced on $\mathcal P_\Theta$ by $(\Theta, g)$.

Here we consider the parametric family of beta distributions, a family of probability measures on $[0,1]$ with density with respect to the Lebesgue measure parameterized by two positive scalars $x, y >0$
\begin{equation*}
p(t;x,y) = \frac{\Gamma(x+y)}{\Gamma(x)\Gamma(y)} t^{x-1}(1-t)^{y-1}, \quad t\in[0,1].
\end{equation*}
We consider the Riemannian manifold composed of the parameter space $\Theta=\mathbb{R}_+^*\times\mathbb{R}_+^*$ and the Fisher information metric $g$, and by extension denote by beta manifold the pair $(\mathcal B, g)$, where $\mathcal B$ is the family of beta distributions
\begin{equation*}
\mathcal B = \{B(x,y)=p(\cdot;x,y) dt, x>0, y>0\}.
\end{equation*}
Here $dt$ denotes the Lebesgue measure on $[0,1]$. The beta family is a natural exponential one with log-partition function %potential function:
\begin{equation}\label{logpartition}
\varphi (x,y) = \ln\Gamma(x+y) - \ln \Gamma(x) - \ln \Gamma(y).
\end{equation}
and so the Fisher information admits a simple expression:
\begin{equation*}
I(x,y)= - \text{Hess}\,\varphi(x, y).
\end{equation*}
Straightforward computations then yield
\begin{equation}\label{metricmatrix}
I(x,y)=\left[\begin{matrix} 
\psi'(x) -\psi'(x+y) & -\psi'(x+y) \\
-\psi'(x+y) & \psi'(y) -\psi'(x+y)
\end{matrix}\right]
\end{equation}
where $\psi$ denotes the digamma function, i.e. 
\begin{equation}
\psi(x) = \frac{d}{dx}\ln\Gamma(x).
\end{equation}
The matrix form of the Fisher metric $g$ is given by \eqref{metricmatrix}, and its infinitesimal length element is
\begin{equation*}
    ds^2 = \psi'(x)dx^2+\psi'(y)dy^2 -\psi'(x+y)(dx+dy)^2.
\end{equation*}

\subsection{Geodesics and geodesic distance}

The distance between two beta distributions is defined as the geodesic distance associated to the Fisher metric in the parameter space
\begin{equation}
d_F(B(x,y),B(x',y')) = \inf_{\gamma} \int_0^1 \sqrt{g(\dot\gamma(t),\dot\gamma(t))}dt,
\label{eq:dist}
\end{equation}
where the infimum is taken over all paths $\gamma:[0,1]\rightarrow \Theta$ such that $\gamma(0)=(x,y)$ and $\gamma(1)=(x',y')$. To compute this distance in practice, one needs to solve the geodesic equation.
\begin{proposition}\label{prop:geodeq}
The geodesics $t\mapsto(x(t),y(t))$ of the beta parameter space $\Theta$ are solutions of
\begin{equation}
\label{geodeq}
\begin{aligned}
\ddot x + a(x,y) \dot x^2 + b(x,y)\dot x\dot y + c(x,y)\dot y^2 = 0,\\
\ddot y + a(y,x) \dot y^2 + b(y,x)\dot x\dot y + c(y,x)\dot x^2 = 0,
\end{aligned}
\end{equation}
where
\begin{equation*}
\begin{aligned}
a(x,y) &=\frac{\psi''(x)\psi'(y) - \psi''(x)\psi'(x+y) - \psi'(y)\psi''(x+y)}{2d(x,y)},\\
b(x,y) &=-\frac{\psi''(x+y)\psi'(y)}{d(x,y)},\\
c(x,y) &=\frac{\psi''(y)\psi'(x+y) - \psi'(y)\psi''(x+y)}{2d(x,y)},\\
d(x,y) &= \psi'(x)\psi'(y) - \psi'(x+y)(\psi'(x)+\psi'(y)).
\end{aligned}
\end{equation*}
\end{proposition}
%We obtain 
No closed form for the geodesics is known, but they can be computed  numerically by  solving \eqref{geodeq}, see Section. \ref{sec:numerics}. Nonetheless we can notice that, due to the symmetry of the metric with respect to parameters $x$ and $y$, the line of equation $x=y$ is a geodesic, where the parameterization is fixed by the unique geodesic equation given by \eqref{geodeq}. 

\begin{figure}
    \centering
    \includegraphics[width=0.7\linewidth]{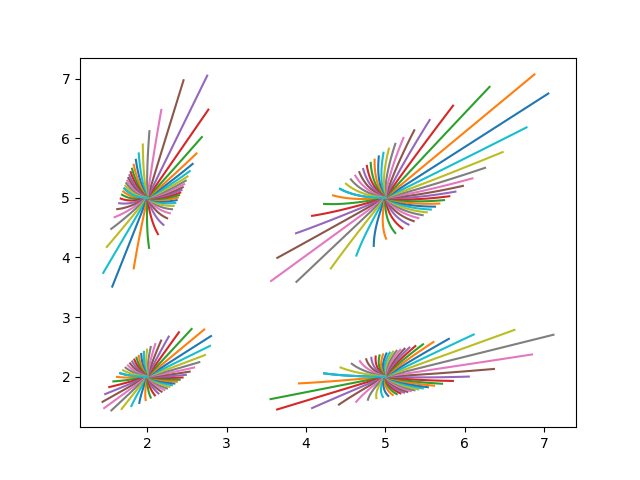}
    \caption{Geodesic balls in the beta manifold.}
    \label{fig:beta_manifold}
\end{figure}

\subsection{Negative curvature and Fréchet mean}

Extension of basic statistical objects such as the mean, the median or the variance to the setting of Riemannian manifolds is now well known; see e.g. \cite{pennec2006intrinsic}. A popular choice to define the notion of mean in a Riemannian manifold is the Fr\'echet mean, also called intrinsic mean, which is defined as the minimizer of the sum of the squared distances to the points that we want to average. In our setting, the intrinsic mean of a set of $\{B_i, i=1,\hdots,n\}$ of beta distributions is given by
\begin{equation*}
    \bar B = \underset{B\in\mathcal B}{\rm{argmin}}\sum_{i=1}^nd_F(B,B_i)^2.
\end{equation*}
The Fr\'echet mean is in general not unique and refers to a set. Uniqueness holds however when the curvature of the Riemannian manifold is negative \citep{karcher1977}, which is the case here.
\begin{theorem}\label{thm:curv}
The beta manifold $(\mathcal B, d_F)$ has negative sectional curvature. %Therefore, any set of beta distributions admits a unique Fréchet mean for the Fisher information geometry.
\end{theorem}
The proof of this theorem, given in the appendix, relies on certain regularity conditions of the polygamma functions, and in particuler on the sub-additivity of the ratio $\psi'/\psi''$ which was proven recently by \cite{yang2017}. 

The space of beta distributions being moreover simply connected and complete as shown in \cite{lebrigant2021}, it is a Hadamard manifold. Therefore any set of beta distributions admits a unique Fréchet mean for the Fisher geometry. This property of the Fisher geometry is an important argument in favor of its use for supervised and unsupervised classification of histograms. Indeed, the popular K-means algorithm, also called Lloyd's algorithm \citep{lloyd1982least}, which seeks to minimize intra-cluster variance, relies on the computation of the mean of the clusters. It can be easily extended to non-linear manifolds by replacing the Euclidean distance and mean by the Riemannian distance and Fréchet mean. 

\subsection{Implementation}\label{sec:numerics}

Geodesics (see Fig. \ref{fig:beta_manifold}) are computed through the exponential and logarithm maps of the Riemannian manifold, which can be implemented by solving respectively the initial and boundary value problems associated to the geodesic equation \eqref{geodeq}. This allows to approximate the distance, by averaging the norm of the velocity of the discretized geodesic between two points.
The Fréchet mean can be found in practice using a gradient descent algorithm commonly referred to as the Karcher flow. It simply consists in applying the following update on the current estimate $\hat B$ of the mean at each iteration:
\begin{equation*}
    \hat B \leftarrow \exp_{\hat B}\left(\frac{\tau}{n}\sum_{i=1}^n\log_{\hat B}(B_i)\right)
\end{equation*}
for some step size $\tau>0$. Here $\exp$ and $\log$ denote the Riemannian exponential and logarithm maps. Our implementation is available in the open-source Python package geomstats \url{http://geomstats.ai}.

\section{Applications to medical data analysis}\label{sec:applications}

In this section with exemplify the theoretical results of the previous section on histograms computed from medical data in two different contexts.

\subsection{Data-sets}

\paragraph*{Cardiac Shape Deformations}
Firstly, we use a data-base of 204 shapes of right ventricles (RV) of the heart extracted from 3d-echocardiographic sequences \citep{moceri2017}. 3d-meshes were extracted from the sequences with semi-automatic software (TomTec 4D RV-Function 2.0) and post-processed to compute markers of deformation during the cardiac cycle. We focus on the systole, that is the period during which the RV contracts to eject blood through the pulmonary valve. As a marker of deformation, we use the Area Strain (AS), that represents the local stretching of the RV surface. This feature is of growing interest among clinicians to assess cardiac function. It was shown to be a predictor of survival in pulmonary hypertension \citep{moceri2017}.

\begin{figure}
    \centering
    \includegraphics[width=0.52\textwidth]{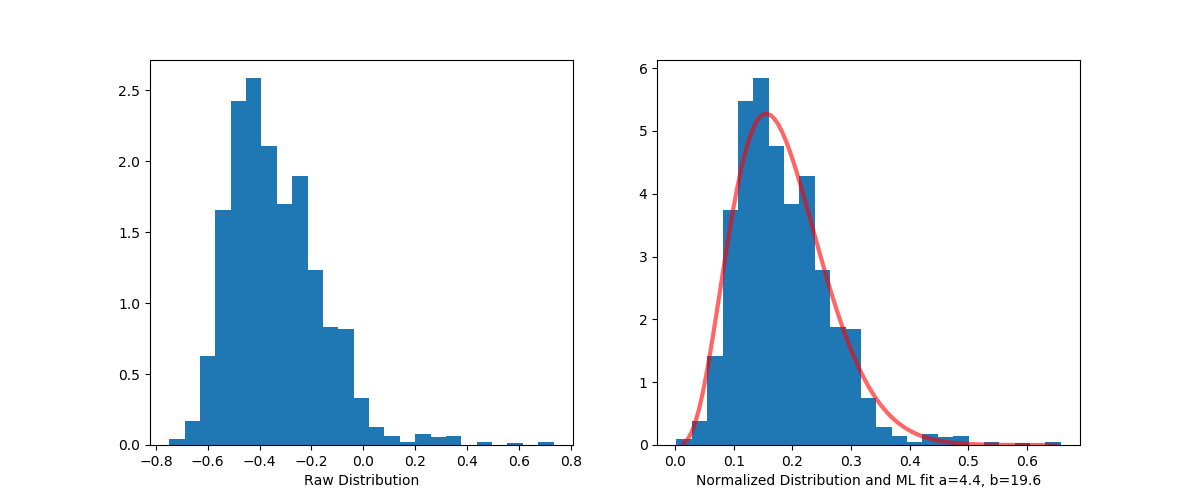}
    \caption{Example of the normalization and ML representation of a subject's AS distributions in the space of beta distributions}
    \label{fig:histograms}
\end{figure}

\begin{figure}
    \centering
    \includegraphics[width=0.7\linewidth]{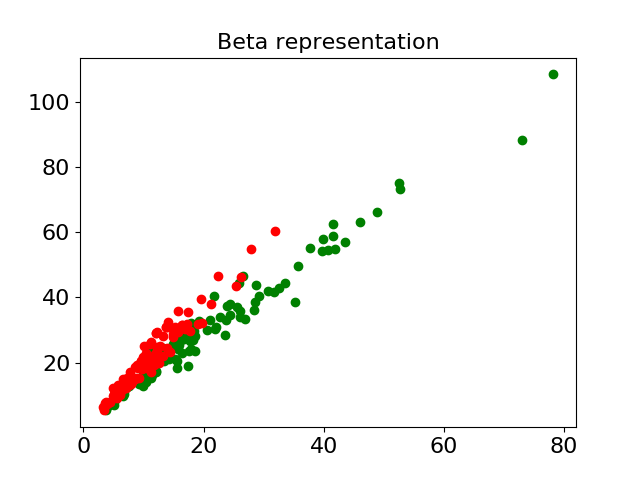}
    \includegraphics[width=0.7\linewidth]{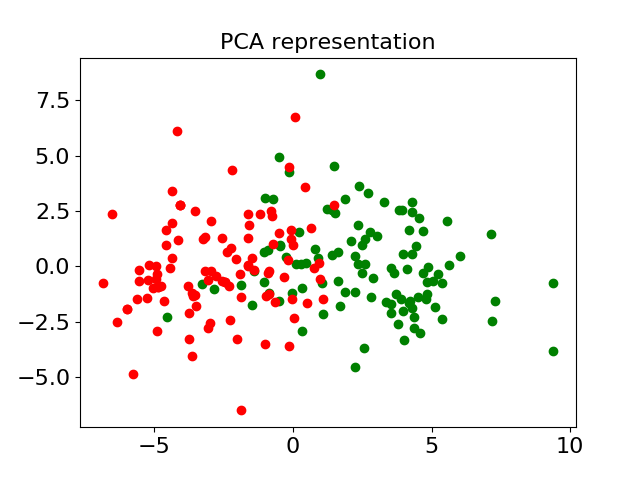}
    \caption{AS data represented in the parameter space of beta distributions (up) and in the space of two first principal components (down).}
    \label{fig:AS}
\end{figure}

For each subject, the RV at time $t$ is represented by a 3d-mesh of 822 vertices and 1587 triangular cells. Note that all vertices and cells correspond to one-another across subjects, but this may not be the case in most applications. This allows us to compare statistics computed in the original data space and in the space of beta distributions. We compute the relative area change of each cell $k$ between the end of diastole ($t_0$) and the end of systole ($t_1$):
\begin{equation}
    AS_k = \frac{a_k^{t_1} - a_k^{t_0}}{a_k^{t_0}},
\end{equation}
where $a_k^t$ is the area of the triangle $k$ at time $t$ for a given mesh. This results in a distribution of values greater than~$-1$. Due to the tissue's nature, the values should in fact be bounded by constants $p<q$. However, the acquisition noise causes some values to lie outside $[p,q]$ although physically infeasible. We thus normalize the data to lie in the interval $[0, 1]$ by applying the transformation 
\[x \mapsto \frac{\min( \max(x, p), q)-p}{q-p}.\]
Finally, we map each patient's histogram to the space of beta distributions by estimating the parameters $x,y$ by maximum likelihood (ML). Fig. \ref{fig:histograms} represents the data for a given patient with the estimated probability density function and Fig. \ref{fig:AS} the obtained representation of the population's histograms in the manifold of beta distributions.

\paragraph*{Cortical Thickness maps}

The measure of brain atrophy is a crucial tool in the analysis of neurodegenerative diseases. In particular, Cortical Thickness (CTh) measured by structural Medical Resonance Imaging (MRI) has been proposed as a biomarker for the early diagnosis of Alzheimer’s disease \citep{pini2016brain,busovaca2016alzheimer} and for the prediction of conversion from Mild Cognitive Impairment to Alzheimer's Disease \citep{Wei2016, Sun2017}. In many of these studies, central tendency measures such as the mean or the median are used to represent the biomarker, either in the whole cortex, in regions of interest or in the voxels, leading to an important loss of information. More recent ones used histogram analysis \citep{giulietti2018whole, ruiz2018alzheimer}, and in \citep{rebbah2019geometry}, the authors use the Fisher geometry of generalized gamma distributions to perform this analysis. However, cortical thickness is a bounded quantity, and therefore beta distributions seem to be more natural candidates to model their distribution. Moreover, the negative curvature of the beta geometry makes it a suitable candidate in clustering procedures based on computation of cluster means, such as the K-mean algorithm used here.

%Brain cortical thickness is an anatomical measure of common use when dealing with neurodegenerative diseases.

The data used in this application were extracted from MR scans selected from the Alzheimer’s Disease Neuroimaging Initiative (ADNI) database\footnote{\url{http://adni.loni.usc.edu/about/}}. Indeed, the initial subjects were not age- and sex-matched and our procedure consisted in randomly selecting subjects. In addition, some of the subjects were excluded because of a low diagnosis reliability (according to ADNI criteria) and others because of unsuccessful CTh measurement due to poor image quality. The resulting population is composed of 143 subjects: 71 healthy controls subjects and 72 Alzheimer’s disease patients.

CTh was measured from 3D T1-weighted MR images using the Matlab toolbox CorThiZon \citep{querbes2009}. There is no unique definition of CTh. In \citep{macdonald2000} it is obtained as the distance between uniquely associated pairs of points on bounding surfaces. Uncoupled methods do not rely on a given set of associated landmarks but create them using techniques coming from shape matching \citep{tagare1997}. Finally, a quite different category of algorithms makes use of an elliptic partial differential equation to obtain a field of deformation between the boundaries \citep{yezzy2003}. The CorThiZon toolbox, developed at Inserm, implements a Laplace equation based method that falls within this last category. 

For each subject, the data consist in measurements of CTh along the whole cortical ribbon. The overall procedure is the one described in \citep{querbes2009}, with a normalized voxel size of 1mm along all directions. Due to the variability of the size and shape of the brain among the population of study, we obtain samples of unequal length. This results in the lack of a common representation space for the data, a problem that is solved by the histogram-based approach that we propose here. Indeed, just as for the AS data, we compute for each patient the histogram of CTh measurements, from which we estimate the parameters of a beta distribution by ML. Histograms are normalized beforehand with respect to the maximal CTh value among the population.

\begin{figure}
    \centering
    \includegraphics[width=0.7\linewidth]{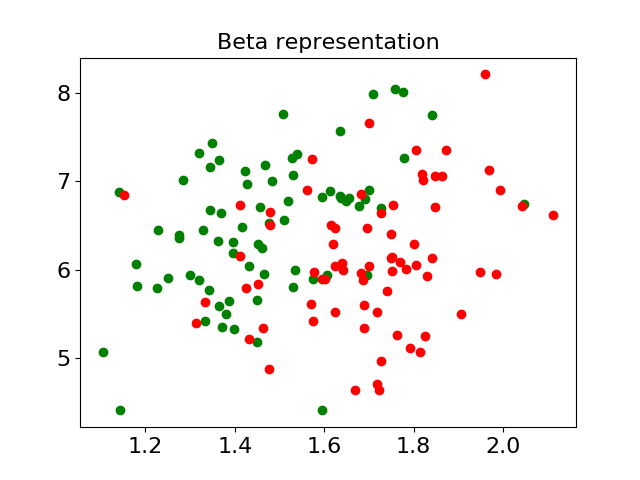}
    \caption{CTh data represented in the parameter space of beta distributions.}
    \label{fig:CW}
\end{figure}

\subsection{Methodology}

In both cases, the subjects are divided in two classes: diseased and controls. In order to assess whether the proposed representation and geometry are relevant, we perform supervised and unsupervised classification in the Riemannian manifold of beta distributions. 
% To use the Riemannian structure, all the distances are computed by solving the problem \ref{eq:dist}. 
We compare the results both when using the Riemannian metric and the Euclidean metric on the parameter space $(R_*^+ )^2$ . We chose to use the K-nearest- neighbor (KNN) and K-means algorithms as they only require computation of distances on non-linear manifolds. Both supervised (SKM) and unsupervised (UKM) versions of K-means are compared.
Finally, in the case of the AS data, as we have cell-to-cell correspondence, we can run the algorithms in the original data space R1587 to assess the loss of information when considering only two-parameter distributions instead of the whole histograms. We also compare our method with a standard dimensionality reduction performed by principal component analysis (PCA). This cannot be done for the CTh data since the mapping between values and brain regions differs across patients.
All models are trained and tested in a 5-fold cross-validation fashion, and we assess the classification accuracy in all cases. We used the scikit-learn package \citep{scikit-learn} to perform these experiments.
%In both cases, the subjects are divided in two classes: diseased and controls. In order to assess whether the proposed representation and geometry are relevant, we perform supervised and unsupervised classification in three different representation spaces: (1) the Euclidean space of the original data, (2) the space spanned by the first two principal components obtained by PCA and (3) the space of parameters of the beta distributions fitted to the data-point histograms. Both the Euclidean and the Fisher information geometries are considered in the space of beta parameters. For the CTh data-set, only this last representation space is used as the original data-vectors all have different dimensions.

\subsection{Results}

\begin{figure}
    \centering
    \includegraphics[width=0.90\linewidth]{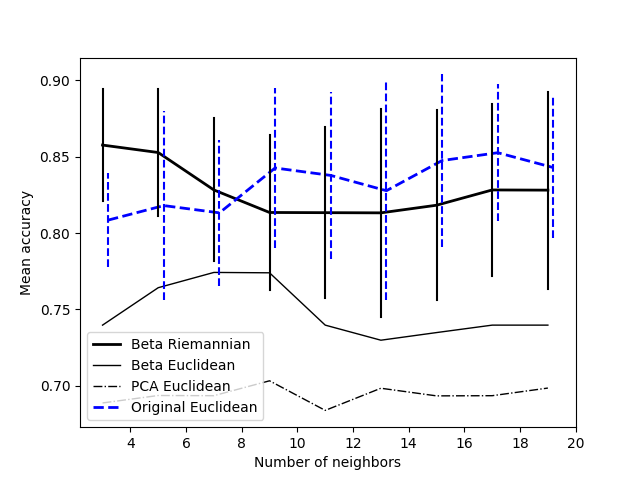}
    \caption{Mean accuracy of KNN classification on the AS data over 5-fold cross validation. Vertical segments indicate standard deviation for the original Euclidean and the beta Riemannian representations.}
    \label{fig:AS_knn}
\end{figure}

\begin{figure}
    \centering
    \includegraphics[width=0.90\linewidth]{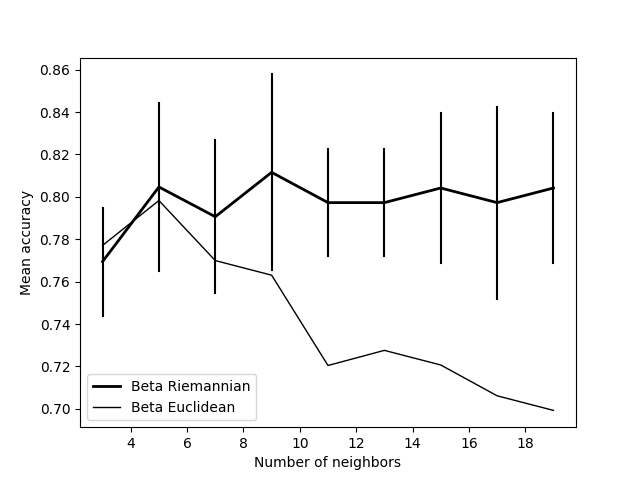}
    \caption{Mean accuracy of KNN classification on the CTh data over 5-fold cross validation. Vertical segments indicate standard deviation for the beta Riemannian representation.}
    \label{fig:CW_knn}
\end{figure}

\paragraph*{Supervised classification}
Tables \ref{tb:AS_classif} and \ref{tb:CTh_classif} show the mean classification accuracy over 5-fold cross validation for each setting: (1) the space of parameters of the beta distributions fitted to the data-point histograms and, when applicable, (2) the Euclidean space of the original data, and (3) the space spanned by the first two principal components obtained by PCA. The number of neighbors for the KNN algorithm is chosen to be $K=7$. This choice is compared to other values through 5-fold cross validation, %on a training test (
see Fig. \ref{fig:AS_knn} and \ref{fig:CW_knn}.
%) coupled with an evaluation on the corresponding test set (Table \ref{fig:AS_knn_test}). 

\begin{table}[hb]
\begin{center}
\caption{Mean (and standard deviation) of the classification accuracy for the AS data on 5-fold cross-validation}\label{tb:AS_classif}
%\begin{tabular}{ccccc}
%\hline
% & \multicolumn{2}{c}{Beta} & Original & PCA  \\
% & Euclidean & Riemannian & Euclidean & Euclidean  \\
% \hline
%KNN & 0.77 (0.09) & {\bf 0.83} (0.05) & 0.81 (0.05) & 0.67 (0.32) \\
%SKM & 0.66 (0.06) & 0.81 (0.06) & {\bf 0.85} (0.03) & 0.72 (0.28) \\
%\hline
%\end{tabular}
\begin{tabular}{ccccc}
\hline
 & Original & PCA & \multicolumn{2}{c}{Beta} \\
 & Euclidean & Euclidean & Euclidean & Riemannian \\
 \hline
KNN & 0.81 (0.05) & 0.67 (0.32)& 0.77 (0.09) & {\bf 0.83} (0.05) \\
SKM & {\bf 0.85} (0.03) & 0.72 (0.28) & 0.66 (0.06) & 0.81 (0.06)\\
\hline
\end{tabular}
\end{center}
\end{table}

\begin{table}[hb]
\begin{center}
\caption{Mean (and standard deviation) of the classification accuracy for the CTh data on 5-fold cross-validation}\label{tb:CTh_classif}
\begin{tabular}{ccc}
\hline
 & \multicolumn{2}{c}{Beta} \\
 & Euclidean & Riemannian \\
 \hline
KNN & 0.77 (0.05) & {\bf0.79} (0.04) \\
SKM & 0.66 (0.10) & {\bf0.80} (0.05)\\
\hline
\end{tabular}
\end{center}
\end{table}

These results show that in the space of beta parameters, the Fisher information geometry performs significantly better than Euclidean geometry. Moreover, this performance is comparable to that of classification in the original data space, despite the considerable reduction of dimension (from 1587 to 2). This is illustrated in Fig. \ref{fig:AS_knn}, where both methods perform alternatively better as the number of neighbors changes. 

\begin{figure}
    \centering
    \includegraphics[width=0.90\linewidth]{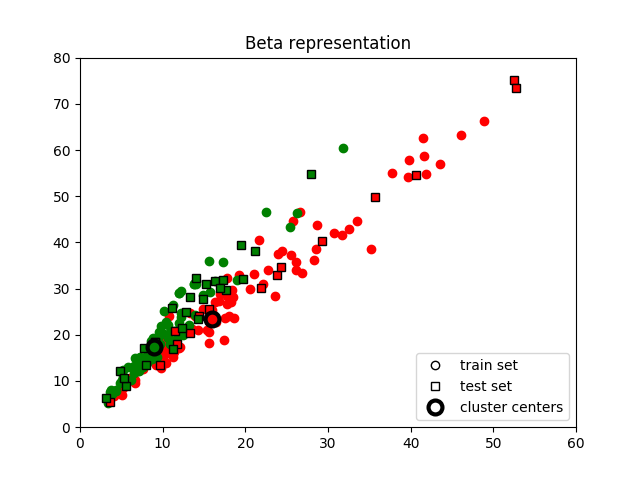}
    \includegraphics[width=0.90\linewidth]{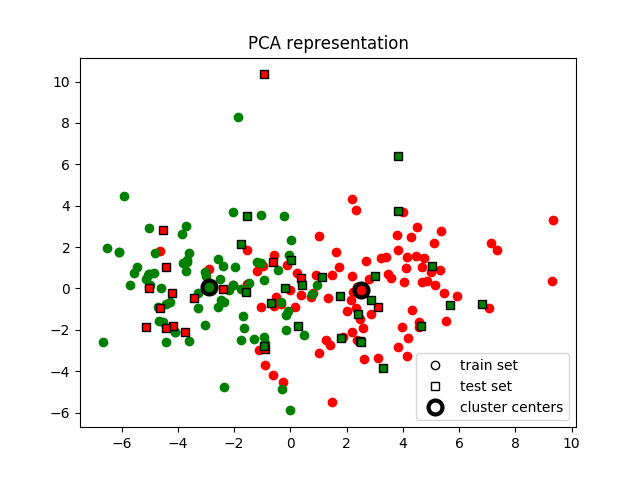}
    \caption{Supervised K-means on the AS strain data in beta and PCA representations.}
    \label{fig:AS_skm}
\end{figure}

In comparison, the two-dimensional representation space provided by PCA performs significantly worst on average. This can be explained by the fact that the performance of dimensionality reduction of PCA depends on the choice of training set, and certain train/test splittings lead to poor performance. This is illustrated in Fig. \ref{fig:AS_skm}: in one of the cross-validation splittings, the test set for each class in the PCA representation is closer to the center of the other class computed from the training test. This has the effect of lowering the average accuracy over the whole cross-validation process. This problem does not occur in the two-dimensional beta representation, as the mapping from one data-point to a pair of parameters $(x,y)$ does not depend on the other points present in the data-set.

\paragraph*{Unsupervised classification}
The clustering performance of the $K$-means algorithm is shown for the same settings in Tables \ref{tb:AS_clust} and \ref{tb:CTh_clust}. These results confirm that the Fisher information geometry in the space of beta parameters is more suited to compare histograms than Euclidean geometry. Unlike in the supervised setting, the two-dimensional representation given by PCA performs well, since in this case the whole data-set was used to perform the dimensionality reduction.

\begin{table}[hb]
\begin{center}
\caption{Clustering accuracy for the AS data}\label{tb:AS_clust}
\begin{tabular}{ccccc}
\hline
 & \multicolumn{2}{c}{Beta} & Original & PCA  \\
 & Euclidean & Riemannian & Euclidean & Euclidean \\
 \hline
UKM & 0.60 & 0.80 & {\bf 0.84} & {\bf 0.84} \\
\hline
\end{tabular}
\end{center}
\end{table}

\begin{table}[hb]
\begin{center}
\caption{Clustering accuracy for the CTh data}\label{tb:CTh_clust}
\begin{tabular}{ccc}
\hline
 & \multicolumn{2}{c}{Beta} \\
 & Euclidean & Riemannian \\
 \hline
UKM & 0.61 & {\bf 0.82} \\
\hline
\end{tabular}
\end{center}
\end{table}

\section{Conclusion}

In this work, we have studied the geometry of the beta manifold and shown in particular that it was negatively curved. We have illustrated its use as a non linear representation space for histograms of medical data, which presents the advantage of being low-dimensional and can easily be visualized. This common representation is also particularly helpful when the number of measurements vary from one subject to another. The examples studied here tend to show that the Fisher information metric is a more adapted choice of metric on this space than the Euclidean one to compare and classify histograms. Finally, the dimensionality reduction performed in the process does not result in a significant loss in performance. Moreover, unlike a standard method like PCA, the quality of the dimensionality reduction does not rely on the choice and size of the data-set: each data-point is mapped to an element of the beta manifold independently of the others. This makes the beta representation particularly interesting for small data-sets. Comparison with other geometries, such as that of the gamma and generalized gamma manifolds, and exploration of other supervised and unsupervised classification algorithms will be the object of future work.

\begin{ack}
The authors would like to warmly thank Pamela Moceri (CHU Nice) and Nicolas Duchateau (Universit\'e Lyon 1) for acquiring and curating the heart data. N. Guigui has received funding from the European Research Council (ERC) under the European Union’s Horizon 2020 research and innovation program (grant agreement G-Statistics No 786854). This research has been conducted within the FP2M federation (CNRS FR 2036)
\end{ack}

\bibliographystyle{ifacconf}
\bibliography{references}    % bib file to produce the bibliography with bibtex (preferred)

\appendix
\section{Proof of Proposition \ref{prop:geodeq}}    

The geodesic equations are given by
\begin{equation*}
\begin{aligned}
\ddot x + \Gamma_{xx}^x\dot x^2 + 2\Gamma_{xy}^x\dot x\dot y + \Gamma_{yy}^x \dot y^2=0\\
\ddot y + \Gamma_{xx}^y\dot x^2 + 2\Gamma_{xy}^y\dot x\dot y + \Gamma_{yy}^y \dot y^2=0
\end{aligned}
\end{equation*}
where the $\Gamma_{ij}^k$'s denote the Christoffel symbols of the second kind. These can be obtained from the Christoffel symbols of the first kind $\Gamma_{ijk}$ and the coefficients $g^{ij}$ of the inverse of the metric matrix
\begin{equation*}
\Gamma_{ij}^k = \Gamma_{ijl} g^{kl}.
\end{equation*}
Here we have used the Einstein summation convention. Since the Fisher metric is a Hessian metric (see \cite{totaro2004curvature}), the Christoffel symbols of the first kind can be obtained as 
\begin{equation*}
\Gamma_{ijk} = \frac{1}{2}\varphi_{ijk},
\end{equation*}
where $\varphi$ is the log-partition function \eqref{logpartition} and lower indices denote partial derivatives. Straightforward computation yields the desired equations.

\section{Proof of Theorem \ref{thm:curv}}

The sectional curvature of a Hessian metric is given by
\begin{equation*}
K = \frac{1}{4(\det g)^2} R_{xyxy}
\end{equation*}
where 
\begin{equation*}
\begin{aligned}
R_{xyxy} &= -\varphi_{yy} (\varphi_{xxx}\varphi_{xyy} - \varphi_{xxy}^2)\\
&+ \varphi_{xy} (\varphi_{xxx}\varphi_{yyy} - \varphi_{xxy}\varphi_{xyy})\\
&+ \varphi_{xx} (\varphi_{xxy}\varphi_{yyy} - \varphi_{xyy}^2).
\end{aligned}
\end{equation*}
Here lower indices denote partial derivatives with respect to the corresponding variables, and $\varphi$ is the log-partition function \eqref{logpartition}. Their computation yields
\begin{align*}
\varphi_{xxx} &= \psi''(x) - \psi''(x+y),  \\
\varphi_{yyy} &= \psi''(y) - \psi''(x + y), \\
\varphi_{xxy} &= \varphi_{xyy} = -\psi''(x+y),
\end{align*}
and the determinant of the metric is given by
\begin{equation*}
\det g =d(x,y)= \psi'(x)\psi'(y) - \psi'(x+y)(\psi'(x)+\psi'(y)).
\end{equation*}
This gives
\begin{equation*}
\begin{aligned}
K(x,y) = (\psi''(x+y)(\psi'(x)\psi''(y)+\psi''(x)\psi'(y))\\
- \psi'(x+y)\psi''(x)\psi''(y))/(4d(x,y)^2).
\end{aligned}
\end{equation*}
Factorizing the numerator by $\psi''(x)\psi''(y)\psi''(x+y)$ yields
\begin{equation*}
\begin{aligned}
K(x,y) = \frac{\psi''(x)\psi''(y)\psi''(x+y)}{4\,d(x, y)^2}(F(x)+F(y)-F(x+y)),%\frac{\psi'(x)}{\psi''(x)} &+ \frac{\psi'(y)}{\psi''(y)} \\
%& - \frac{\psi'(x+y)}{\psi''(x+y)}\bigg).
\end{aligned}
\end{equation*}
where $F=\psi'/\psi''$. Since $\psi''$ is negative, the first factor is negative. Moreover, it has been shown in \cite[Corollary 4]{yang2017} that the function $F$ is sub-additive. This means that the second factor is positive, yielding the desired result.

\end{document}